\documentclass[sigconf,screen]{acmart}

\AtBeginDocument{%
  }

\copyrightyear{2024}
\acmYear{2024}
\setcopyright{rightsretained}
\acmConference[HAI '24]{International Conference on Human-Agent
Interaction}{November 24--27, 2024}{Swansea, United Kingdom}
\acmBooktitle{International Conference on Human-Agent Interaction (HAI '24),
November 24--27, 2024, Swansea, United Kingdom}
\acmDOI{10.1145/3687272.3688324}
\acmISBN{979-8-4007-1178-7/24/11}

\usepackage{acmart-taps}
\usepackage{mathtools} 
\usepackage{dsfont} 
\usepackage{diagbox}
\usepackage{tabularx}
\newcommand{\argmin}{\operatornamewithlimits{argmin}} 
\usepackage{algpseudocode}
\usepackage{algorithm}

\begin{document}

\title[Semifactual Explanations for Reinforcement Learning]{Semifactual Explanations for Reinforcement Learning}

\author{Jasmina Gajcin}
\email{gajcinj@tcd.ie}
\orcid{0000-0002-8731-1236}
\affiliation{%
  \institution{Trinity College Dublin}
  \streetaddress{42A Pearse Street}
  \city{Dublin}
  \country{Ireland}
  \postcode{D02R123}
}

\author{Jovan Jeromela}
\email{jeromelj@tcd.ie}
\orcid{0000-0003-3272-9114}
\affiliation{%
  \institution{Trinity College Dublin}
  \streetaddress{25 Westland Row}
  \city{Dublin}
  \country{Ireland}
  \postcode{D02W272}
}

\author{Ivana Dusparic}
\email{Ivana.Dusparic@tcd.ie}
\orcid{0000-0003-0621-5400}
\affiliation{%
  \institution{Trinity College Dublin}
  \streetaddress{42A Pearse Street}
  \city{Dublin}
  \country{Ireland}
  \postcode{D02R123}
}

\begin{abstract}

    Reinforcement Learning (RL) is a learning paradigm in which the agent learns from its environment through trial and error. Deep reinforcement learning (DRL) algorithms represent the agent's
    policies using neural networks, making their decisions difficult to interpret. Explaining the behaviour of DRL agents is necessary to advance user trust, increase engagement, and facilitate integration with real-life tasks. Semifactual explanations aim to explain an outcome by providing ``even if'' scenarios, such as ``even if the car were moving twice as slowly, it would still have to swerve to avoid crashing''. Semifactuals help users understand the effects of different factors on the outcome and support the optimisation of resources. While extensively studied in psychology and even utilised in supervised learning, semifactuals have not been used to explain the decisions of RL systems. In this work, we develop a first approach to generating semifactual explanations for RL agents. We start by defining five properties of desirable semifactual explanations in RL and then introducing SGRL-Rewind and SGRL-Advance, the first algorithms for generating semifactual explanations in RL. We evaluate the algorithms in two standard RL environments and find that they generate semifactuals that are easier to reach, represent the agent's policy better, and are more diverse compared to baselines. Lastly, we conduct and analyse a user study to assess the participant's perception of semifactual explanations of the agent's actions.

\end{abstract}



\maketitle

\section{Introduction}
\label{secIntro}

Reinforcement learning (RL) is a machine learning paradigm in which the agent learns a policy from environment interactions through a trial-and-error process \citep{sutton20018reinforcement}. In deep reinforcement learning (DRL), the RL  learning process is powered by deep neural networks, and DRL algorithms have found applications in numerous domains, including robotics, healthcare, and navigation \citep{arulkumaran2017deepRL,hasegawa2022advantage,bhattacharya2023reinforcement}. However, neural networks are black-box algorithms, meaning their decisions are difficult to interpret. This lack of transparency is a serious concern, as it can affect the user's trust and RL deployment, especially in high-risk domains \citep{puiutta2020explainable,jeromela2023onboarding}. Prior works have also indicated that allowing the user to interpret AI systems increases user acceptance of AI systems \cite{maehigashi2023experimental} and enables effective collaboration between humans and AI agents \cite{jang2023structured}.  Furthermore, there are often legal requirements for the users of AI systems to be presented with an explanation \citep{celar2023how,goodman2017european}. This has led to the proliferation of the field of eXplainable Artificial Intelligence (XAI). Challenges of XAI include devising algorithms to generate explanations for black-box models as well as evaluating different types of explanations \citep{rosenfeld2019explainability,jeromela2023onboarding}.

Counterfactual explanations have been established as a powerful explanation technique in XAI \citep{wachter2017counterfactual,guidotti2022counterfactual}. To explain an outcome, counterfactuals present the user with an alternative scenario where the outcome would have been different (e.g. ``\textit{If only your loan application was for a lower amount, it would have been accepted}''). Counterfactual explanations are considered to be user-friendly, as they are actionable, contrasting and inherent to human reasoning \citep{olson2021counterfactual,huber2023ganterfactual,molnar2020interpretable}.

Semifactual explanations are similar to counterfactuals in that they also aim to explain an outcome by exploring alternative worlds. However, unlike counterfactuals, which explore worlds where the outcome has changed, semifactuals present those alternatives where the outcome remains the same \cite{mccloy2002semifactual}.  For example, when explaining why a loan application was rejected, a semifactual might state that ``\textit{Even if your salary was twice as high, the application would still have been rejected}''. While counterfactual explanations can be used to understand how a negative outcome can be avoided, semifactuals could be utilised to optimise resources necessary for a positive outcome. For example, an AI system utilising semifactuals could explain to the farmer that even if they used less fertiliser, the yield would have remained the same \citep{aryal2023even}. \citet{mccloy2002semifactual} reported that semifactuals engage psychological processes different from those triggered by counterfactuals. First, semifactuals focus on confirming the outcome even in the alternative scenario. They make the outcome seem more correct and immutable, making them more convincing. Second, semifactuals seem to elicit less negative emotions in users when it comes to understanding negative outcomes such as loan rejection. While counterfactuals might potentially be perceived as blaming the users for the outcome, semifactuals reinstate the outcome and reconfirm that nothing could have been done to prevent it \cite{mccloy2002semifactual}. Research into semifactuals as an XAI technique is fairly new, and while there are methods for generating semifactuals in supervised learning tasks \citep{aryal2023even,aryal2024semi,kenny2021generating,kenny2024utility}, they have not yet been explored for RL. Additionally, while their benefits have been speculated, only a fraction of works conducted user studies to explore how users perceive semifactuals and how they affect the interaction between humans and RL-based agents \citep{aryal2023even}. 

In this work, we explore how semifactuals can be utilised to explain the decisions of RL agents. First, we propose a set of desired properties that informative semifactual explanations should meet. These properties are based on the sequential and stochastic nature of RL algorithms. Next, we provide two algorithms\footnote{The code and evaluation details at: \hyperlink{https://github.com/anonymous902109/SGRL}{https://github.com/anonymous902109/SGRL}.} for searching for semifactuals in RL: SGRL-Rewind\footnote{Semifactuals Generator for Reinforcement Learning (SGRL).} that searches for changes in the past that would have maintained an outcome; and SGRL-Advance that looks into the future to explore how future actions might maintain an outcome. We evaluated our algorithms in two RL tasks against S-GEN, a state-of-the-art approach for generating semifactuals for supervised learning \cite{kenny2024utility}.  Our findings show that using our algorithms generates semifactual explanations that are more true to the model (have better fidelity) and are more diverse. The analysis of our subsequent user study indicates that semifactuals may positively affect user understanding of RL-led decisions, although further research remains needed to better investigate the effects on task performance. 
Our contributions are as follows:
\begin{itemize}
    \item We propose a set of five semifactual properties that determine the most informative semifactual.
    \item We introduce SGRL-Advance and SGRL-Rewind, two algorithms for semifactual search which optimise the semifactual properties.
    \item We evaluate our approach against a state-of-the-art baseline in two RL environments and perform a user study to assess the effect of semifactual explanations on user understanding of RL agents.
\end{itemize}

\section{Related Work}
\label{secSotA}

Counterfactuals have been explored in RL both to explain decisions and to correct behaviour. \citet{olson2021counterfactual} published the first approach to generating counterfactuals in RL based on generative modelling. Similarly, \citet{huber2023ganterfactual} proposed GANterfactual-RL, 
a model-agnostic generative approach based on the StarGAN architecture for generating counterfactuals. \citet{gajcin2024raccer} introduced RACCER, an approach which relies on RL-specific counterfactual properties and searches for the counterfactual instance within the agent's execution tree. Additionally, counterfactual explanations providing sequences of actions that can be used to avoid a negative outcome have been explored \citep{gajcin2024acter,tsirtsis2021counterfactual,tsirtsis2023finding}.

Semifactual explanations have only recently been introduced as an XAI technique \citep{aryal2024even}. Approaches to generating semifactuals in supervised learning can be divided into counterfactual-guided methods, which require a counterfactual to be generated to guide the search for the semifactual, and counterfactual-free methods, which typically rely on some distance metric to choose the furthest instance of the same class as the query \citep{aryal2024even}. In counterfactual-guided methods, semifactuals are often searched for between the original instance and the decision boundary. \citet{kenny2021generating} introduced PIECE, an approach to generating plausible counterfactuals in image-based tasks by identifying and resetting them to their expected values for the counterfactual class. A counterfactual-guided semifactual generation can follow the same process, except that it terminates before the decision boundary is reached \cite{kenny2021generating}. On the other hand, counterfactual-free methods often rely on a metric to choose a semifactual that is the farthest away from the original instance while retaining the same outcome or classification. For example, \citet{nugent2009gaining} proposed generating a neighbourhood of instances around the original instance and using logistic regression to classify them. The instance from the neighbourhood that belongs to the same class as the original instance but has the lowest probability of such a classification is then chosen as the semifactual explanation. Thus, the approach reaffirms the classification decision by showcasing similar, yet more extreme, examples within the same class. \citet{aryal2023even} propose the MDN (Most Distant Neighbour) approach, which searches for the semifactual as an instance that differs most significantly from the original in a chosen key feature. Notably, \citet{kenny2024utility} introduce S-GEN, a method for generating semifactuals that rely on causality. To generate semifactuals, authors start by creating a structural causal model (SCM) around the query instance. Neighbouring instances represent nodes, while actions that can transform one state into another are edges. Achieving the semifactual means taking action in the SCM. S-GEN then searches for semifactuals by optimising three metrics -- \textit{gain}, \textit{robustness}, and \textit{diversity}. The \textit{gain} is determined by the distance to the states, taking into account a specified user preference. \textit{Robustness} is measured in relation to instances in the neighbourhood of the semifactual in the same class. The \textit{diversity} encourages the generation of semifactuals with diverse features to enable personalisation.

While explored in supervised learning, to the best of our knowledge, there are no works on generating semifactual explanations for RL. In this work, we explore how semifactuals can be defined for RL tasks, 
and propose the first algorithm for generating semifactuals. 

\section{Semifactual Generation for RL}
\label{secApproach}

In this section, we introduce SGRL-Advance and SGRL-Rewind, two algorithms for generating semifactual explanations in RL. First, Section \ref{secPreliminaries} covers the preliminary considerations. Section \ref{secProperties} defines the five properties that we put forward as optimisation criteria for the selection of the most informative RL semifactual explanations. Then, in subsection \ref{secAlgorithm}, we describe the two algorithms that generate semifactual explanations based on these properties.

\begin{figure*}
    \centering
    \includegraphics[width=0.8\linewidth]{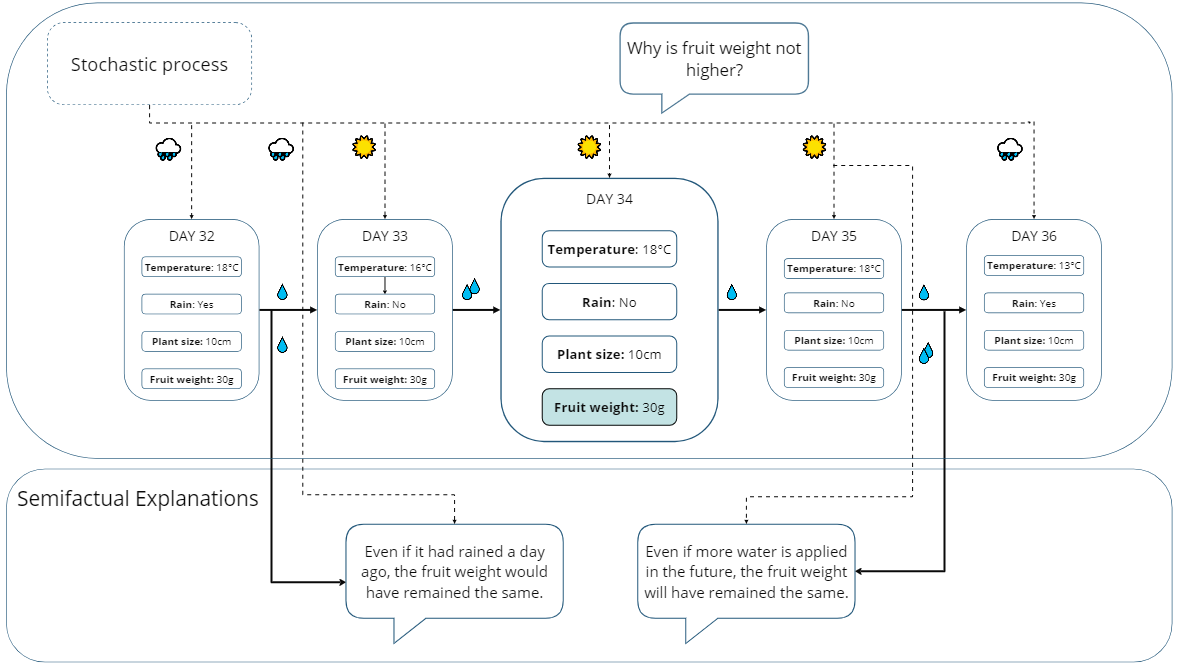}
 \caption{Forward and backward semifactuals: In an agricultural task, semifactuals can be used to explain why the yielded fruit weight at a specific time was not higher. The backward semifactual looks into the past and reports that more rain would not have changed the fruit weight. Conversely, the forward semifactual looks into the future and states that more watering in the future is not going to change the outcome. Dashed lines represent the effect of stochastic processes on the environment's state. In this case, a backward semifactual relies on a non-actionable change of weather, while the forward one explains the outcome through an actionable change. }
    \label{figSemifactualGeneration}
\end{figure*}

\subsection{Preliminaries}
\label{secPreliminaries}

\subsubsection{Outcomes in Semifactual Explanations}

The goal of semifactuals is to explain \textit{the outcome} of an RL policy. We use a broad definition of outcome in this work, referring to any function of the state. 

\begin{definition}[Outcome]
  Let $\mathcal{F}$ be the function whose properties semifactual explanations aim to illustrate. Then, the outcome of a state $s$ is the outcome of the function $\mathcal{F}$ in that state $s$:
  \begin{equation}
    O(s) := \mathcal{F}(s)
\end{equation}
\end{definition}

In RL, there is a wide array of outcomes of interest. For example, semifactuals can be used to explain an agent's action choice, high or low rewards, or an agent's choice to follow a specific objective or goal. Our approach can be used to explain any such outcome that can be defined as a function of the agent's state. The outcomes can be manually defined based on the user's interests or automatically inferred from the policy, such as achieving a low or high reward in the state. 

\subsubsection{Stochastic Configurations}

In a typical RL task, the agent's action policy is shaped by the rewards the agent receives after performing actions in the environment. Both the function determining the rewards as well as the agent's policy may be stochastic \cite{sutton20018reinforcement}. Therefore, generating semifactual explanations requires reasoning about the stochastic processes in the environment, which can affect the agent's decisions. To reason formally about stochastic processes, we define \textit{stochastic context} $\mathcal{P}$ to be the set of all stochastic processes $P^i$ that the agent cannot control.

\begin{equation}
    \mathcal{P} := \bigcup_{i \in I} P^i
\end{equation}

We then define a \textit{stochastic configuration} $\mathcal{P}_T$ as the stochastic context along a particular trajectory $T$.

\begin{equation}
    \mathcal{P}_T := \bigcup_{i \in I} P_T^i
\end{equation}

To illustrate these definitions, let us reconsider the example of the farmer wondering how the watering of the plants affected the yield. Factors such as daily precipitation or atmospheric temperature are among factors that cannot be influenced by the farmer and are thus elements of the stochastic context, $\mathcal{P}$. An example of a trajectory $T$ would be the past days leading to the day on which the outcome (the harvest yield) is determined. The amounts of rain and measured temperatures during these days would then be in the stochastic configuration $\mathcal{P}_T$. Determining the actual values in $\mathcal{P}_T$  allows looking into the past states in $T$ and identifying factors that can be changed, thus enabling the generation of informative semifactuals. Such an approach is consistent with the previously proposed definition of stochastic configurations for counterfactuals \cite{gajcin2024acter}. In this work too, we consider that the stochastic configuration $\mathcal{P}_T$ can be known in full for the trajectory of interest $T$. 
 
\subsubsection{Temporality of Semifactual Explanations}

Unlike supervised learning, RL deals with sequential execution. For this reason, semifactual statements can be phrased in two ways: either by looking into the past or by looking into the future. For example, consider a farmer wondering about their yield projection. A semifactual on how the past affects outcome might explain that even if they had applied more fertiliser in the past weeks, the yield would have stayed the same. Alternatively, a semifactual focusing on a future scenario might state that even if they applied more fertiliser in the upcoming days, the yield would not have changed. We thus define two types of semifactual explanations: \textit{backward semifactuals} that explore alternative scenarios in the past that retain the same outcome and \textit{forward semifactuals} that look into how future scenarios might have retained the outcome, as illustrated in Figure \ref{figSemifactualGeneration}.

\begin{definition}
\label{defBSmethods}
    Let $(s_{n-k},$ $a_{n-k}),\dots$, $(s_{n-1},$ $ a_{n-1})$ be an arbitrary number of immediate prior state-action pairs that led the agent to the factual state $s_n$ and the outcome $\mathcal{O}(s_n)$ to be explained.
    
    The \textit{backward method} of searching for semifactuals generates alternative state-action pairs that would also result in the same outcome as that of the factual state: $(s'_{n-k},$ $a'_{n-k}), \dots,$ $(s'_{n-1},$ $a'_{n-1})$. All such $s'_{n}$ states are \textit{backward semifactuals}.

    The \textit{forward method} of searching for semifactuals begins at the factual state $s_n$ and then generates alternative future scenarios that retain the outcome of the factual state: $(s'_{n+1}$,$ $$a'_{n+1}), \dots,$ $(s'_{n+k},$ $ a'_{n+k})$. All alternative future states $s'_{n+k}$ are \textit{forward semifactuals}.
\end{definition}

Forward and backward semifactuals are conflated in supervised learning due to a one-step prediction process that does not assume a temporal connection between instances. In RL, however, we have to consider both ways to explain the event. \citet{mccloy2002semifactual} explored a similar distinction in psychology research in counterfactual explanations and has found that different cognitive processes are engaged when using previous and future causes to explain an outcome. The role of temporality in counterfactual explanations in RL was also discussed previously \cite{gajcin2024redefining}.

\subsubsection{Actionability of Semifactual Explanations}

Semifactual explanations require changing the original state. Changes to a RL state can occur from two sources -- agent's actions or stochastic processes in the environment. While the first results in actionable changes, the second is 
non-actionable. For example, while an RL agent can choose to increase the amount of fertiliser in a farming task, the agent cannot control if it rains. Thus, semifactual explanations might differ from the factual state in the amount of fertiliser added, in the weather ``observed'', or in both the weather and fertiliser.

Both actionable and non-actionable semifactuals have a role in explaining outcomes in RL. Actionable semifactuals can help the users understand how actions affect outcomes, while non-actionable semifactuals could explain how stochastic processes in the environment affect the outcome. In this work, we allow both types of changes and see them as equal when looking for semifactuals.
However, further research is needed into how these different types of semifactuals affect user understanding of RL policies. 

\subsection{Desired Semifactual Properties for RL}
\label{secProperties}

Finding a semifactual explanation for the given state $s$ necessitates finding an alternative state $s'$ so that the outcome is the same in both states. Since there might be infinitely many such $s'$ states, we ought to define properties of desirable semifactual states that lead to an informative explanation.

\subsubsection{Validity}
\label{secValidity}

Validity denotes the criterion that the outcome in the factual state $s$ and the semifactual state $s'$ is the same. Formally:

\begin{equation}
    V(s,s') := 
\begin{cases}
    1 & \text{if } \mathcal{O}(s) = \mathcal{O}(s') \\
    0 & \text{otherwise}
\end{cases}
\end{equation}

The validity of semifactuals is analogous to the validity of counterfactuals \cite{verma2020counterfactual,gajcin2024redefining}, for which the outcome must be different from (rather than equal to) the outcome of the factual state. In this work, as we consider action choices as outcomes, we only consider a semifactual state $s'$ valid if the same action is predicted in $s'$ as in $s$ by the agent's policy $\pi$.

\subsubsection{Temporal Distance}
\label{secSparsity}

Semifactuals should take place in the same temporal context of the agent's execution as the state being explained. Intuitively, telling a farmer that more watering of the fields 10 years ago would not have increased the yield is not as informative as indicating that more watering last month would not have changed the outcome. We define temporal distance as the distance in terms of RL actions between the original and the semifactual state. For a state $s$ being explained, a semifactual $s'$, and a sequence of actions $A$ between the two states, temporal distance is defined as:

\begin{equation}
    TD(s, s', A) :=  len(A)
\end{equation}

The action sequence $A$ can go backward or forward from the factual state, depending on whether we are considering backward or forward semifactuals. This is formalised in Definition \ref{defBSmethods} and illustrated in Figure \ref{figSemifactualGeneration}.

\subsubsection{Stochastic Uncertainty}
\label{secStochastic}

To be informative, a semifactual explanation must offer a stronger argument than the state in question. Coming back to our loan example from the beginning of the paper, saying ``even if your income was \textit{twice as high}, your application would still be rejected'' is more telling than stating that the loan would have been rejected if the applicant's income was lower than it actually is. For this reason, desired semifactual explanations should prioritise paths with a higher possibility of an outcome change, as they lead the search to the part of the state space where the outcome is more likely to change. We call this property \textit{stochastic uncertainty}, in line with a similar property for counterfactuals \cite{gajcin2024acter}. The property is defined for the factual state $s$ and the sequence of actions $A'$ leading to the semifactual state. Formally, we have:

\begin{equation}
    SU(s,A') := \mathds{P}(\pi(A'(s_{n+k}|w)) =\pi(s)  \ \forall w \in W) 
\end{equation}

where $W$ is the set of all possible stochastic configurations and $\pi(s)$ is the action that the black-box policy chooses in state $s$.

\subsubsection{Fidelity}
\label{secFidelity}

In order to ensure the semifactuals are representative of the policy they are to clarify, we rely on the \textit{fidelity}, as defined for counterfactuals by \citet{gajcin2024raccer}. Fidelity measures the probability that the sequence of actions $A$ (leading to some semifactual state) would be chosen by the black-box policy from the given (factual) state $s$. 

\begin{align}
    F(s,A) := & 1 - \prod_{a \in A} \mathrm{softmax}(Q(s,\mathcal{A}))[a] \\
    =& 1 - \prod_{a \in A} \frac{\exp{(Q(s,a))}}{\sum_{a'\in\mathcal{A}}\exp{(Q(s,a'))}}  \notag
\end{align}

$Q(s,a)$ denotes the policy-dependent Q-value, that is, the valuation of taking the action $a$ from state $s$. While in this case, we use the softmax function over all possible actions from state $s$, other methods of normalising the likelihood of taking action $a$ is possible. If we consider the example of a farmer, an explanation that would indicate that the yield would have been the same even if he applied additional fertiliser a day before the harvest would have low fidelity, as doing so is an unreasonable practice and would have a corresponding low Q-value. Thus, fidelity is a mechanism of hindering semifactuals that are not in line with the policy to be explained.

\subsubsection{Exceptionality}
\label{secExceptionality}

As proposed by \citet{aryal2023even}, desired semifactuals are supposed to be surprising or even counterintuitive. For example, knowing that the yield would have been the same even if the field suffered a drought might reassure the farmer in their decisions and reinforce their trust in the system. Thus, informative semifactuals ought to be \textit{exceptional}, i.e. we want the chosen explanation to prioritise paths where the unexpected happens. We formalise this property as a measure of the likelihood of the path of $k-1$ state-action pairs leading from the factual state $s=s_1$ to the semifactual state $s'=s_k$:

\begin{equation}
    E(s_1, s_k, A) := \sum_{i \in \{1,\dotsc,k-1\}}  \mathds{P}(s_{i+1} | s_i, a_i)
\end{equation}

For computational reasons we use the likelihood (the sum of probabilities) rather than cumulative probability. Intuitively, exceptionality prioritises paths where unexpected things happen. 

\subsection{Proposed NSGA-II-Based Algorithms for Generating Semifactual Explanations}
\label{secAlgorithm}

To find a semifactual that optimises the five desired properties, we define two search algorithms, SGRL-Rewind and SGRL-Advance. Both algorithms are model-agnostic and counterfactual-free, i.e. they are applicable to all RL models and do not require searching for counterfactuals first.
SGRL-Rewind and SGRL-Advance search for the semifactual state by exploring the neighbourhood around the original instance. While SGRL-Advance searches for semifactual states that can be reached from the original state in $k$ actions, SGRL-Rewind searches for semifactuals that arise from the agent choosing different actions in the previous $k$ actions. Parameter $k$ denotes the maximum distance between the original and semifactual state.

SGRL-Rewind and SGRL-Advance utilise an evolutionary algorithm NSGA-II (Nondominated Sorting Genetic Algorithm II) \citep{deb2002fast} to generate and evaluate action sequences leading to semifactual states. An overview of the NSGA-II, as used by SGRL-Advance and SGRL-Rewind, is shown in Algorithm \ref{algNSGA-II}.
Through $G$ generations, NSGA-II modifies the initial population of solutions $P_0$ while maintaining a collection of the most promising solutions. For each generation $g$, a child population $Q_{g}$ is obtained from the parent population $P_{g}$ through selection, recombination, and mutation (Line $3$). $P_g$ and $Q_g$ are joined into $R_g = P_g + Q_g$ (Line $4$) and used to obtain the new parent population $P_{g+1}$. Firstly, $R_g$ is split into fronts $F_0, \dots, F_t$ using nondominated sorting where $F_1$ contains the best solutions according to the optimisation objectives (Line $5$). $P_{g+1}$ is populated using solutions from the best fronts until the size $N$ is reached (Lines $7-9$). The last front $F_l$ that can be included is sorted according to the crowding distance (Lines $10 - 13$) and the best solutions are included into $P_{g+1}$ until the size $N$ is reached (Lines $14 - 19$). Using crowding distance promotes solutions from different parts of the solution space and ensures diversity. For both SGRL-Rewind and SGRL-Advance, we use NSGA-II with four properties -- temporal distance, stochastic uncertainty, fidelity, and exceptionally -- as optimisation objectives and the validity property as a constraint.

\begin{algorithm}[t]
\caption{The main loop of the NSGA-II \cite{deb2002fast} adapted for the SGRL-Advance and SGRL-Rewind algorithms}\label{algNSGA-II}
\begin{algorithmic}[1]

\Statex \textbf{Parameters:} Horizon $k$; Population Size $N$; Number of Generations $G$

\Statex \textbf{Input:} Initial population $P_0$ 
\Statex \textbf{Constraints:} $\mathcal{C}=V(s_n, s')$ \Comment{Validity function}
\Statex \textbf{Objectives:} $\mathcal{O}$ = $\{TD(s_n, s', A)$, $1-SU(s_n,A)$,  $F(s_n,A)$, $1-E(s_n,s',A)\}$
 \Comment{Optimisation properties}

\Statex \textbf{Begin:}

\State $P_i = P_0$; $g=0$

\While{$g < G$}
\State $Q_i = \text{get\_child\_population}(P_i)$ \Comment{}{ selection, combination, mutation}
\State $R_{g} = P_{g} + Q_{g}$ \Comment{Combine the parent and child populations}
\State $(F_0 , F_1 ,\dots, F_t)$ $=$ nondominated\_sorting$(R_g , \mathcal{C}, \mathcal{O})$ 
\State $l = 0$; $P_{g+1} = \{\}$

\While{$|P_{g+1}|+|F_l|<N$}
\State $P_{g+1} = P_{g+1} < \cup F_l$   \Comment{Add solutions ordered descendingly into $P_{g+1}$}
\EndWhile

\If{$|P_{g+1}|<PS$}
\State $C_{\text{dist}} = $ crowding\_distance$(F_{l},C_{\text{dist}})$ \Comment{Calculate the distance for the latest n.d. set}
\State \(F'_l =\) sort\_descending\((F_{l},C_{\text{dist}})\) \Comment{Sort based on the crowding distance}
\EndIf

\ForAll{$f \in F'_l$}
\State $P_{i+1} = P_{i+1} \cup f$ \Comment{Add the best solutions}

\If{$|P_{g+1}| \geq N$}
\State \textbf{break}
\EndIf

\EndFor

\EndWhile

\end{algorithmic}
\end{algorithm}

SGRL-Advance explores semifactuals that can be reached by the agent's future actions from the original state. To that end, SGRL-Advance uses NSGA-II to generate and evaluate different action sequences that lead to potential semifactual states. For each action sequence $A^{g, p}, p \in [0, N]$, from generation $g \in [0, G]$ considered by the NSGA-II algorithm, we execute $A^{g, p}$ starting in state $s_n$ and collect all states reached by this rollout as a set $SF$ of potential semifactual states. We store only those states that satisfy validity. Each state in $SF$ is then evaluated according to the four semifactual properties -- fidelity, temporal distance, stochastic uncertainty and exceptionalism. Additionally, the action sequence $A^{g, p}$ is assigned the average value for each property evaluated for $SF(A^{g, p})$. This value is the fitness function value for $A^{g, p}$ and is used to further guide the selection process of NSGA-II. After $G$ generations, a Pareto front of all potential semifactual states is selected from $SF$, and these are presented as the results. Formally, to explain state $s_n$, SGRL-Advance aims to find a state $s'$ that optimises the following:

\begin{align}    
    &\argmin_{s' = A(s_n)} \biggr[ TD(s_n, s', A), 1-SU(s_n, A), \label{forAdvanceGoalsJas} \\ 
    & F(s_n, A),1-E(s_n,s', A) \biggr] \text{ and } V(s_n, s') = 1  \notag
\end{align}

Conversely, SGRL-Rewind uses the backward method of searching for semifactuals. Given a sequence of $k$ previous state-action pairs $[(s_{n-k}, a_{n-k})$, $(s_{n-k+1}, a_{n-k+1}),\dots, (s_{n-1},a_{n-1})]$ leading to the factual state $s_n$, SGRL-Rewind explores semifactual states that could be reached if the agent had chosen an alternative sequence of actions starting in $s_{n-k}$. Formally, NSGA-Rewind optimises the following when explaining state $s_n$: 

\begin{align}    
    &\argmin_{s' = A(s_{n-k})} \biggr[ TD(s_n, s', A), 1 - SU(s_n, A), \label{forRewindGoalsJas} \\ 
    & F(s_n, A),1-E(s_n,s', A) \biggr] \text{ and } V(s_n, s') = 1  \notag
\end{align}



\section{Experimental Evaluation}
\label{secEvaluation}

In this section, we outline our SGRL-Rewind and SGRL-Advance evaluation approach. The evaluation tasks are introduced in Section \ref{secEE}, and the baselines are described in Section \ref{secBaselines}.

\begin{table}
    \centering
    \caption{Parameters used for training a DQN black-box policy $\pi$ in the Stochastic Gridworld and Frozen Lake environments.}
    \label{tabDQNparameters}
    \begin{tabularx}{\linewidth}{c|cc} \toprule 
            \begin{imageonly}\backslashbox{Parameter}{Task}\end{imageonly} & Stochastic Gridworld  & Frozen Lake \\ \midrule
            Number of layers & 2 & 2\\
            Nodes in each layer & [256, 256] & [256, 256]\\
            Learning rate & $5 \cdot 10^{-4}$ & $5 \cdot 10^{-4}$ \\
            Training time steps ($N$) & $3 \cdot 10^{5}$& $3 \cdot 10^{5}$ \\
        \bottomrule    
        \end{tabularx}
\end{table}

\subsection{Evaluation Environments}
\label{secEE}

We evaluate SGRL-Rewind and SGRL-Advance in two RL tasks.

\subsubsection{Stochastic Gridworld}

In the Stochastic Gridworld task, the agent navigates a 5x5 grid world. At each step, the agent chooses between 6 actions -- UP, DOWN, LEFT, RIGHT (to make one step in that direction), CHOP (to chop down an obstacle if located next to it), and SHOOT (to kill the dragon). Agent's goal is to shoot the dragon by performing a SHOOT action. This action is only effective if both the agent and dragon are in the same column or the same row and there are no obstacles between them. The obstacles in the environment include trees and walls, which the agent can tear down or navigate around to reach the dragon. Tearing down a tree is less expensive than tearing down a wall. Trees can also spontaneously regrow, and walls can be rebuilt by processes outside of the agent's control, making the environment stochastic.

\subsubsection{Frozen Lake}

In the Frozen Lake task, the agent aims to reach the goal state of a 5x5 grid world. At each step, the agent can choose between moving in any of the four directions or choosing to exit the game. Exiting the game when the agent has reached the goal square is met with a positive reward; otherwise, it does not have any effect on the agent's position. Some squares in the grid world are frozen, so there is a chance of the agent moving in an unintended direction from them.

\begin{table}[!t]
\centering
\caption{Parameters used for generating semifactual explanations using SGRL-Advance and SGRL-Rewind in the Stochastic Gridworld and Frozen Lake environments. The same parameters are used for both approaches.}
\label{tabSGRLparameters}
        \begin{tabularx}{\linewidth}{c|cc} \toprule                 \begin{imageonly}\backslashbox{Parameter}{Task}\end{imageonly} & Stochastic Gridw.   & Frozen Lake \\ \midrule
                Number of generations ($G$)  & 25 & 25 \\
                Population size ($N$) & 24 & 24 \\
                Horizon ($k$) & 3 & 3\\
                Evaluation dataset size ($|D|$) & 600 & 500 \\   
             \bottomrule    
        \end{tabularx}
\end{table}

\subsection{Baselines}
\label{secBaselines}

SGRL-Advance and SGRL-Rewind are the first approaches for generating semifactuals in RL. For this reason, we compare them to methods developed for supervised learning. As a baseline, we use S-GEN \citep{kenny2024utility}, an approach to generating semifactuals in supervised learning. S-GEN generates semifactuals by optimising a loss function consisting of three components: gain, robustness, and diversity. We use the non-causal version of S-GEN, as the causal version requires a manually crafted causal model of the environment. S-GEN can create diverse explanations and requires a diversity parameter, which indicates how many explanations will be generated. We use three baselines: S-GEN1, S-GEN3, and S-GEN5, which generate $1$, $3$, and $5$ diverse semifactuals, respectively \citep{kenny2024utility}. 

We start by training a black-box RL policy $\pi$ in both tasks, which we want to explain. We use a DQN \citep{mnih2013playing} to train $\pi$ in both environments. However, our approach is model-agnostic and can be used to explain any RL policy. The parameters for training $\pi$ in both environments are given in Table \ref{tabDQNparameters}. 
In each environment, we then execute $\pi$ and collect factual states to explain. For each possible action $a'$ in the environment, we collect $100$ states in which action $a'$ was not chosen. Our goal is then to generate semifactual states answering the question of why the agent did not choose $a'$ in the collected states. In total, we collect $100$ states for each action in both environments, resulting in $600$ factual states in the Stochastic Gridworld and $500$ in the Frozen Lake environment. For each factual state, we then generate semifactual states according to our approaches SGRL-Advance and SGRL-Rewind. Additionally, we create semifactuals using baseline approaches S-GEN1, S-GEN3 and S-GEN5. Table \ref{tabSGRLparameters} shows the parameters for the baseline approaches.

\section{Results}
\label{secResults}

In this section, we present the evaluation results. Firstly, we evaluate SGRL-Advance and SGRL-Rewind against the baselines on semifactual properties (Section \ref{secSFPropertiesEvaluation}). In Section \ref{secFeatureGain} we evaluate feature gain and in Section \ref{secDiversity} diversity of generated semifactuals. Section \ref{secUserStudy} details the conditions and results of our user study. 

\begin{table*}[t]
    \centering
    \caption{The average values of semifactual properties for semifactuals generated using SGRL-Advance, SGRL-Rewind, and the three baselines in the Stochastic Gridworld and Frozen Lake environments.}
\fontsize{8.5}{10.5}\selectfont
    \begin{tabularx}{\linewidth}{c|ccccc|ccccc} \toprule 
        Task & \multicolumn{5}{c|}{Stochastic Gridworld} & \multicolumn{5}{c}{Frozen Lake} \\ \midrule
        \begin{imageonly}\backslashbox{Metric}{Approach}\end{imageonly} & SGRL-Ad. & SGRL-Re. & S-GEN1 & S-GEN3 & S-GEN5 & SGRL-Ad. & SGRL-Re. & S-GEN1 & S-GEN3 & S-GEN5 \\ \midrule 
        Generated semifactuals (\%) & \textbf{86.66} & \textbf{86.66} & 83.3 & 83.33 & 83.33 & 100 & 100 & 80 & 80 & 80\\ 
        Validity ($=1$) & \textbf{1.0 }& \textbf{1.0 }& \textbf{1.0} & \textbf{1.0} &\textbf{ 1.0} & \textbf{1.0} & \textbf{1.0} & \textbf{1.0} & \textbf{1.0} & \textbf{1.0} \\ 
        Temporal distance ($\downarrow$) & 0.91 & \textbf{0.87} & 0.95 & 0.96 & 0.96 & 0.92 & \textbf{0.88} & 0.99 & 0.99 & 0.99
        \\ 
        Fidelity ($\downarrow$) & \textbf{0.13} & 0.24 & 0.27 & 0.27 & 0.27 & \textbf{0.14} & 0.17 & 0.97 & 0.97 & 0.97 \\ 
         Stochastic uncertainty ($\downarrow$) & 0.80 & 0.84 &\textbf{0.46} & \textbf{046}& 0.47 & \textbf{0.86} & 0.87 & 0.98 & 0.98 & 0.98 \\ 
        Exceptionality ($\downarrow$) & \textbf{0.76} & 0.83 & 0.92 & 0.92 & 0.92 &  \textbf{0.16} & 0.29 & 0.97 & 0.97 & 0.97\\ 
        \bottomrule    
    \end{tabularx}
    \label{tabResults}
\end{table*}

\subsection{Semifactual Properties Evaluation}
\label{secSFPropertiesEvaluation}

Our first goal is to evaluate how semifactuals perform on the five RL-specific properties defined in Section \ref{secProperties} -- validity, temporal distance, stochastic uncertainty, fidelity, and exceptionalism. Additionally, we record the percentage of factual states for which a semifactual was successfully generated. As some properties need a sequence of actions leading to the semifactual, they cannot be straightforwardly evaluated for the S-GEN approach, which can only generate a semifactual state without the sequence of actions leading to it. For this reason, for each semifactual generated by S-GEN, we search for a sequence of actions that could lead to the semifactual using an evolutionary algorithm and use this sequence of actions to evaluate the semifactual properties. 

The results are shown in Table \ref{tabResults}. In both environments, SGRL-Advance and SGRL-Rewind find the semifactual states for more factual states than the baselines. The average validity is always $1$, as this property was set as a constraint, and all candidate states not satisfying it were rejected. SGRL-Rewind performs best on the temporal distance metric in both environments, generating semifactual states that are closest in the number of actions to the original state. On the other hand, SGRL-Advance performs best on the metrics of fidelity and exceptionalism in both environments. Thus, SGRL-Advance-generated semifactuals are most representative of the policy being explained and they showcase most unexpected events in which the agent does not change its action choice. Stochastic uncertainty is the only property where the baseline outperforms SGRL-Advance and SGRL-Rewind algorithms. Specifically, SGEN1 and SGEN3 yielded better results for stochastic uncertainty in the Stochastic Gridworld task. In the Frozen Lake task, however, SGRL-Advance achieved the best results. Given that the SGRL algorithms outperformed S-GEN in all but one case across the two environments indicates that the resulting semifactuals are easier to reach, represent the underlying policy better, and are more illustrative of even how unexpected scenarios still maintain the outcome. As the first RL-specific semifactual generation approaches, we think that these results are promising, and we encourage future work in the area to use the SGRL algorithms as their baseline.

\begin{table*}[t]
    \centering
    \caption{The gain and diversity evaluation for semifactual explanations generated using SGRL-Advance and SGRL-Rewind and the baseline approaches in the Stochastic Gridworld and Frozen Lake tasks.}
    \begin{tabularx}{\linewidth}{c|ccccc|ccccc} \toprule 
        Task & \multicolumn{5}{c|}{Stochastic Gridworld} & \multicolumn{5}{c}{Frozen Lake} \\ \midrule
        \begin{imageonly}\backslashbox{Metric}{Approach}\end{imageonly} & SGRL-Ad. & SGRL-Re. & S-GEN1 & S-GEN3 & S-GEN5 & SGRL-Ad. & SGRL-Re. & S-GEN1 & S-GEN3 & S-GEN5\\ \midrule
        Gain ($\uparrow$)  & 4.92 & \textbf{8.37} & 6.67 & 6.71 & 6.65 & 5.23 & 8.07 & \textbf{12.04} & 12.02 & 12.02 \\ 
        Diversity ($\uparrow$) & 6.11 & \textbf{6.80} & 0 & 4.71 & 4.95 & \textbf{6.46} & 6.12 & 0.0 & 0.0 & 0.0 \\ 
        \bottomrule    
    \end{tabularx}
    \label{tabResultGain}
\end{table*}

\subsection{Feature Gain}
\label{secFeatureGain}

Semifactual explanations can be used to enhance decision-making and reduce user effort without changing the outcome. For example, a semifactual can advise the user that even if they used less water, the yield would have remained the same, thus saving resources. For this reason, our experiments measured \textit{gain}
. Intuitively, it measures the difference between the factual and semifactual instances; larger differences can indicate higher resource savings. Formally, the gain was defined as feature distance metric by \citet{kenny2024utility}:

\begin{equation}
    G(x, x') = |x - x'|_2
\end{equation}

As shown in Table \ref{tabResultGain}, in the Stochastic Gridworld, SGRL-Rewind achieves the biggest gain. In the Frozen Lake, however, the biggest difference in features is reported by the S-GEN1 algorithm. This result is not surprising, given that gain is one of the optimisation goals for S-GEN but not for SGRL algorithms. Unlike in supervised learning, features alone may not indicate the gain in RL. Namely, states in RL can have different features and still be very close to each other in execution. For this reason, we do not use this property as an optimisation goal for either SGRL-Advance or SGRL-Rewind.

\subsection{Diversity}
\label{secDiversity}

Diversity has been recognised as an important characteristic of semifactual explanations, ensuring that they can be presented to users with different preferences \cite{kenny2024utility}. For example, while one user might find it more beneficial to maintain yield while saving water, another might want to achieve the same result while spending less fertiliser. For each factual state $x$, we evaluate the diversity of a set of generated semifactual $X'$ as: 

\begin{equation}
    D(x, X') = \frac{1}{|X'|} \sum_{x' \in X'} |x - x'|_2
\end{equation}

The results are shown in Table \ref{tabResultGain}. Similar as in the case of gain, our SGRL algorithms do not optimise for diversity, but the S-GEN algorithms do. Still, in both environments, our approaches generate more diverse sets of explanations. In the Stochastic Gridworld, SGRL-Rewind performs best according to the diversity metric, while in the Frozen Lake task, SGRL-Advance achieves the highest diversity. The high diversity of explanations generated by SGRL-Advance and SGRL-Rewind is a consequence of the underlying use of the NSGA-II algorithm which generates a set of diverse solutions. 

\subsection{User Study}
\label{secUserStudy}

\begin{figure}
    \centering
    \includegraphics[width=\linewidth]{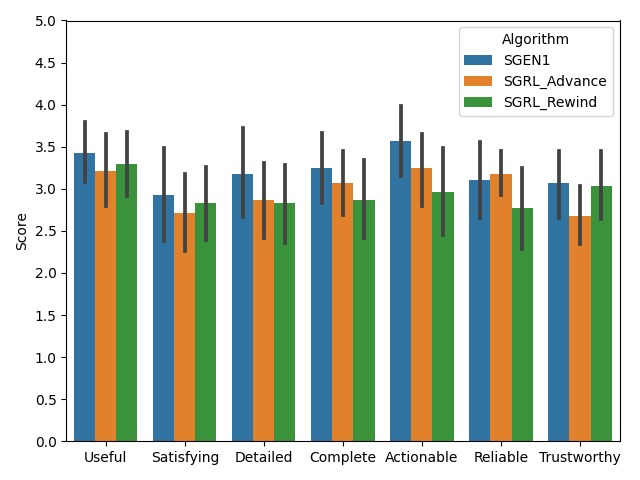}
    \caption{User satisfaction scores for semifactuals generated by S-GEN1, SGRL-Advance and SGRL-Rewind based on explanations goodness metrics \cite{hoffman2018metrics}.}
    \label{figUserStudyResults}
\end{figure}

We conducted a user study to evaluate how different types of semifactual explanations are perceived by users and to assess if semifactual explanations help users understand and predict the behaviour of RL agents. Specifically, we compared explanations generated by SGRL-Advance, SGRL-Rewind, and S-GEN1. We conducted the study using examples from a Stochastic Gridworld task, as this task is easy to understand and requires no prior knowledge.

We recruited $90$ participants from English-speaking countries (US, UK, Canada, Australia, Ireland, and New Zealand) through the Prolific platform. All participants were reimbursed for their time according to the Prolific payment policy. We divided the participants into 3 groups. The first group received explanations generated by SGRL-Advance, the second explanations by SGRL-Rewind, and the third by SGEN1. Before the study, participants were introduced to the task and semifactual explanations. The study was divided into 3 parts -- training, testing, and user satisfaction survey. In the training part, users were shown examples of RL states and semifactual explanations showing why the agent did not choose actions CHOP and SHOOT in that state. We limited our examples to these two actions to reduce cognitive load, as they are most informative. Additionally, to reduce cognitive load, we offer users only one explanation chosen at random from a diverse set of explanations. Each user was shown 9 training examples during the training phase. In the testing phase, users are shown RL states and asked to predict the next action the agent will choose in the state without
any explanations. The states shown to users during the training and testing phase were chosen by the HIGHLIGHTS algorithm \cite{amir2018highlights} to ensure they capture the most important decisions. Finally, in the last stage of the study, the participants were asked to rate explanations on a 1-5 Likert scale based on the explanation goodness metrics \cite{hoffman2018metrics}.

Users who have seen explanations generated by the baseline S-GEN1 were able to correctly predict the agent's actions in $68.62\%$ of cases. However, users who were presented with explanations generated by SGRL-Rewind achieved an accuracy of $75.00\%$, while users who have seen SGRL-Advance were correct in $77.58\%$ of questions. We performed a non-parametric one-tailed Mann-Whitney test and found no significant difference between the explanations. 

Regarding the perceived utility all explanations received goodness scores above $2.5$ (Figure \ref{figUserStudyResults}) with no statistically significant difference between the three algorithms. The fact that the outcome of the SGRL algorithms was not perceived as better than S-GEN1 might indicate that the task was not complex enough to showcase the RL-specific benefits of SGRL. Still, the high overall scores regarding the participants' perception of semifactuals are encouraging.

\section{Limitations and Future Work}
\label{secFutureWork}

Our work is an initial step in enabling semifactuals to explain the decisions of RL agents. Our preliminary user study aligns with the prior findings from other fields \cite{aryal2024semi,mccloy2002semifactual}, that semifactual explanations are liked by users and that they may support human-agent interaction. However, we did not find that users preferred our RL-specific approaches to the baselines adapted from supervised learning tasks. Future studies, potentially involving a qualitative analysis as well, might investigate whether semifactuals of more complex RL tasks would benefit from our algorithms' explicit consideration of agent trajectories and the optimisation properties that are well-defined in RL. On the technical level, we considered that the stochastic configuration $\mathcal{P}_T$ can be known in full for the trajectory of interest $T$. 
Future work should investigate alternative approaches on semifactual generation semifactual generation for offline RL tasks. 

\section{Conclusion}
\label{secConclusion}

In this work, we explored the problem of generating semifactual explanations in RL. We defined five semifactual properties that ensure semifactuals are informative in RL tasks. We introduced two model-agnostic algorithms, SGRL-Advance and SGRL-Rewind, which generate semifactual explanations in RL by optimising these properties. We evaluated the algorithms in two stochastic RL environments and found that they can generate diverse semifactuals that are easier to reach and more representative of the agent's policy compared to a state-of-the-art approach adapted from supervised learning. Lastly, our preliminary user study found an overall positive sentiment towards semifactual explanations of RL tasks. Future studies in more complex RL domains remain necessary.

\begin{acks}
This publication has emanated from research conducted with the financial support of Science Foundation Ireland under Grant number 18/CRT/6223 and SFI Frontiers for the Future grant number 21/FFP-A/8957. 
\end{acks}

\balance
\bibliographystyle{ACM-Reference-Format}
\bibliography{SF4RL}


\begin{thebibliography}{33}


\ifx \showCODEN    \undefined \def \showCODEN     #1{\unskip}     \fi
\ifx \showDOI      \undefined \def \showDOI       #1{#1}\fi
\ifx \showISBNx    \undefined \def \showISBNx     #1{\unskip}     \fi
\ifx \showISBNxiii \undefined \def \showISBNxiii  #1{\unskip}     \fi
\ifx \showISSN     \undefined \def \showISSN      #1{\unskip}     \fi
\ifx \showLCCN     \undefined \def \showLCCN      #1{\unskip}     \fi
\ifx \shownote     \undefined \def \shownote      #1{#1}          \fi
\ifx \showarticletitle \undefined \def \showarticletitle #1{#1}   \fi
\ifx \showURL      \undefined \def \showURL       {\relax}        \fi
\providecommand\bibfield[2]{#2}
\providecommand\bibinfo[2]{#2}
\providecommand\natexlab[1]{#1}
\providecommand\showeprint[2][]{arXiv:#2}

\bibitem[Amir and Amir(2018)]%
        {amir2018highlights}
\bibfield{author}{\bibinfo{person}{Dan Amir} {and} \bibinfo{person}{Ofra Amir}.} \bibinfo{year}{2018}\natexlab{}.
\newblock \showarticletitle{HIGHLIGHTS: Summarizing Agent Behavior to People}. In \bibinfo{booktitle}{\emph{Proceedings of the 17th International Conference on Autonomous Agents and MultiAgent Systems}} (Stockholm, Sweden) \emph{(\bibinfo{series}{AAMAS '18})}. \bibinfo{publisher}{International Foundation for Autonomous Agents and Multiagent Systems}, \bibinfo{address}{Richland, SC}, \bibinfo{pages}{1168–1176}.
\newblock


\bibitem[Arulkumaran et~al\mbox{.}(2017)]%
        {arulkumaran2017deepRL}
\bibfield{author}{\bibinfo{person}{Kai Arulkumaran}, \bibinfo{person}{Marc~Peter Deisenroth}, \bibinfo{person}{Miles Brundage}, {and} \bibinfo{person}{Anil~Anthony Bharath}.} \bibinfo{year}{2017}\natexlab{}.
\newblock \showarticletitle{Deep Reinforcement Learning: A Brief Survey}.
\newblock \bibinfo{journal}{\emph{IEEE Signal Processing Magazine}} \bibinfo{volume}{34}, \bibinfo{number}{6} (\bibinfo{date}{Nov} \bibinfo{year}{2017}), \bibinfo{pages}{26--38}.
\newblock
\showISSN{1558-0792}
\urldef\tempurl%
\url{https://doi.org/10.1109/MSP.2017.2743240}
\showDOI{\tempurl}


\bibitem[Aryal(2024)]%
        {aryal2024semi}
\bibfield{author}{\bibinfo{person}{Saugat Aryal}.} \bibinfo{year}{2024}\natexlab{}.
\newblock \showarticletitle{Semi-factual Explanations in AI}.
\newblock \bibinfo{journal}{\emph{Proceedings of the AAAI Conference on Artificial Intelligence}} \bibinfo{volume}{38}, \bibinfo{number}{21}, \bibinfo{pages}{23379--23380}.
\newblock
\urldef\tempurl%
\url{https://doi.org/10.1609/aaai.v38i21.30390}
\showDOI{\tempurl}


\bibitem[Aryal and Keane(2023)]%
        {aryal2023even}
\bibfield{author}{\bibinfo{person}{Saugat Aryal} {and} \bibinfo{person}{Mark~T. Keane}.} \bibinfo{year}{2023}\natexlab{}.
\newblock \showarticletitle{Even if explanations: prior work, desiderata \& benchmarks for semi-factual XAI}. In \bibinfo{booktitle}{\emph{Proceedings of the Thirty-Second International Joint Conference on Artificial Intelligence}} (Macao, P.R.China) \emph{(\bibinfo{series}{IJCAI '23})}. Article \bibinfo{articleno}{732}, \bibinfo{numpages}{10}~pages.
\newblock
\showISBNx{978-1-956792-03-4}
\urldef\tempurl%
\url{https://doi.org/10.24963/ijcai.2023/732}
\showDOI{\tempurl}


\bibitem[Aryal and Keane(2024)]%
        {aryal2024even}
\bibfield{author}{\bibinfo{person}{Saugat Aryal} {and} \bibinfo{person}{Mark~T Keane}.} \bibinfo{year}{2024}\natexlab{}.
\newblock \showarticletitle{Even-Ifs From If-Onlys: Are the Best Semi-Factual Explanations Found Using Counterfactuals As Guides?}
\newblock \bibinfo{journal}{\emph{arXiv preprint arXiv:2403.00980}} (\bibinfo{year}{2024}).
\newblock


\bibitem[Bhattacharya et~al\mbox{.}(2023)]%
        {bhattacharya2023reinforcement}
\bibfield{author}{\bibinfo{person}{Sukriti Bhattacharya}, \bibinfo{person}{Oussema Gharsallaoui}, \bibinfo{person}{Igor Tchappi}, {and} \bibinfo{person}{Amro Najjar}.} \bibinfo{year}{2023}\natexlab{}.
\newblock \showarticletitle{Reinforcement Learning for Sustainable Mobility: Modeling Pedalcoin, a Gamified Biking Application}. In \bibinfo{booktitle}{\emph{Proceedings of the 11th International Conference on Human-Agent Interaction}} (Gothenburg, Sweden) \emph{(\bibinfo{series}{HAI '23})}. \bibinfo{publisher}{Association for Computing Machinery}, \bibinfo{address}{New York, NY, USA}, \bibinfo{pages}{461–463}.
\newblock
\showISBNx{9798400708244}
\urldef\tempurl%
\url{https://doi.org/10.1145/3623809.3623963}
\showDOI{\tempurl}


\bibitem[Celar and Byrne(2023)]%
        {celar2023how}
\bibfield{author}{\bibinfo{person}{Lenart Celar} {and} \bibinfo{person}{Ruth M.~J. Byrne}.} \bibinfo{year}{2023}\natexlab{}.
\newblock \showarticletitle{How people reason with counterfactual and causal explanations for Artificial Intelligence decisions in familiar and unfamiliar domains}.
\newblock \bibinfo{journal}{\emph{Memory \& Cognition}} \bibinfo{volume}{51}, \bibinfo{number}{7} (\bibinfo{year}{2023}), \bibinfo{pages}{1481--1496}.
\newblock
\showISSN{1532-5946}
\urldef\tempurl%
\url{https://doi.org/10.3758/s13421-023-01407-5}
\showDOI{\tempurl}


\bibitem[Deb et~al\mbox{.}(2002)]%
        {deb2002fast}
\bibfield{author}{\bibinfo{person}{K. Deb}, \bibinfo{person}{A. Pratap}, \bibinfo{person}{S. Agarwal}, {and} \bibinfo{person}{T. Meyarivan}.} \bibinfo{year}{2002}\natexlab{}.
\newblock \showarticletitle{A fast and elitist multiobjective genetic algorithm: NSGA-II}.
\newblock \bibinfo{journal}{\emph{IEEE Transactions on Evolutionary Computation}} \bibinfo{volume}{6}, \bibinfo{number}{2} (\bibinfo{date}{April} \bibinfo{year}{2002}), \bibinfo{pages}{182--197}.
\newblock
\showISSN{1941-0026}
\urldef\tempurl%
\url{https://doi.org/10.1109/4235.996017}
\showDOI{\tempurl}


\bibitem[Gajcin and Dusparic(2024a)]%
        {gajcin2024acter}
\bibfield{author}{\bibinfo{person}{Jasmina Gajcin} {and} \bibinfo{person}{Ivana Dusparic}.} \bibinfo{year}{2024}\natexlab{a}.
\newblock \showarticletitle{ACTER: Diverse and Actionable Counterfactual Sequences for Explaining and Diagnosing RL Policies}.
\newblock \bibinfo{journal}{\emph{arXiv preprint arXiv:2402.06503}} (\bibinfo{year}{2024}).
\newblock


\bibitem[Gajcin and Dusparic(2024b)]%
        {gajcin2024raccer}
\bibfield{author}{\bibinfo{person}{Jasmina Gajcin} {and} \bibinfo{person}{Ivana Dusparic}.} \bibinfo{year}{2024}\natexlab{b}.
\newblock \showarticletitle{RACCER: Towards Reachable and Certain Counterfactual Explanations for Reinforcement Learning}. In \bibinfo{booktitle}{\emph{Proceedings of the 23rd International Conference on Autonomous Agents and Multiagent Systems}} (Auckland, New Zealand) \emph{(\bibinfo{series}{AAMAS '24})}. \bibinfo{publisher}{International Foundation for Autonomous Agents and Multiagent Systems}, \bibinfo{address}{Richland, SC}, \bibinfo{pages}{632–640}.
\newblock
\showISBNx{9798400704864}


\bibitem[Gajcin and Dusparic(2024c)]%
        {gajcin2024redefining}
\bibfield{author}{\bibinfo{person}{Jasmina Gajcin} {and} \bibinfo{person}{Ivana Dusparic}.} \bibinfo{year}{2024}\natexlab{c}.
\newblock \showarticletitle{Redefining Counterfactual Explanations for Reinforcement Learning: Overview, Challenges and Opportunities}.
\newblock \bibinfo{journal}{\emph{ACM Comput. Surv.}} \bibinfo{volume}{56}, \bibinfo{number}{9}, Article \bibinfo{articleno}{219} (\bibinfo{date}{apr} \bibinfo{year}{2024}), \bibinfo{numpages}{33}~pages.
\newblock
\showISSN{0360-0300}
\urldef\tempurl%
\url{https://doi.org/10.1145/3648472}
\showDOI{\tempurl}


\bibitem[Goodman and Flaxman(2017)]%
        {goodman2017european}
\bibfield{author}{\bibinfo{person}{Bryce Goodman} {and} \bibinfo{person}{Seth Flaxman}.} \bibinfo{year}{2017}\natexlab{}.
\newblock \showarticletitle{European Union regulations on algorithmic decision-making and a “right to explanation”}.
\newblock \bibinfo{journal}{\emph{AI magazine}} \bibinfo{volume}{38}, \bibinfo{number}{3} (\bibinfo{year}{2017}), \bibinfo{pages}{50--57}.
\newblock


\bibitem[Guidotti(2022)]%
        {guidotti2022counterfactual}
\bibfield{author}{\bibinfo{person}{Riccardo Guidotti}.} \bibinfo{year}{2022}\natexlab{}.
\newblock \showarticletitle{Counterfactual explanations and how to find them: literature review and benchmarking}.
\newblock \bibinfo{journal}{\emph{Data Mining and Knowledge Discovery}} (\bibinfo{year}{2022}), \bibinfo{pages}{1--55}.
\newblock


\bibitem[Hasegawa et~al\mbox{.}(2022)]%
        {hasegawa2022advantage}
\bibfield{author}{\bibinfo{person}{Rintaro Hasegawa}, \bibinfo{person}{Yosuke Fukuchi}, \bibinfo{person}{Kohei Okuoka}, {and} \bibinfo{person}{Michita Imai}.} \bibinfo{year}{2022}\natexlab{}.
\newblock \showarticletitle{Advantage Mapping: Learning Operation Mapping for User-Preferred Manipulation by Extracting Scenes with Advantage Function}. In \bibinfo{booktitle}{\emph{Proceedings of the 10th International Conference on Human-Agent Interaction}} (Christchurch, New Zealand) \emph{(\bibinfo{series}{HAI '22})}. \bibinfo{publisher}{Association for Computing Machinery}, \bibinfo{address}{New York, NY, USA}, \bibinfo{pages}{95–103}.
\newblock
\showISBNx{9781450393232}
\urldef\tempurl%
\url{https://doi.org/10.1145/3527188.3561917}
\showDOI{\tempurl}


\bibitem[Hoffman et~al\mbox{.}(2018)]%
        {hoffman2018metrics}
\bibfield{author}{\bibinfo{person}{Robert~R Hoffman}, \bibinfo{person}{Shane~T Mueller}, \bibinfo{person}{Gary Klein}, {and} \bibinfo{person}{Jordan Litman}.} \bibinfo{year}{2018}\natexlab{}.
\newblock \showarticletitle{Metrics for explainable AI: Challenges and prospects}.
\newblock \bibinfo{journal}{\emph{arXiv preprint arXiv:1812.04608}} (\bibinfo{year}{2018}).
\newblock


\bibitem[Huber et~al\mbox{.}(2023)]%
        {huber2023ganterfactual}
\bibfield{author}{\bibinfo{person}{Tobias Huber}, \bibinfo{person}{Maximilian Demmler}, \bibinfo{person}{Silvan Mertes}, \bibinfo{person}{Matthew~L. Olson}, {and} \bibinfo{person}{Elisabeth Andr\'{e}}.} \bibinfo{year}{2023}\natexlab{}.
\newblock \showarticletitle{GANterfactual-RL: Understanding Reinforcement Learning Agents' Strategies through Visual Counterfactual Explanations}.
\newblock  (\bibinfo{year}{2023}), \bibinfo{pages}{1097–1106}.
\newblock
\showISBNx{9781450394321}


\bibitem[Jang et~al\mbox{.}(2023)]%
        {jang2023structured}
\bibfield{author}{\bibinfo{person}{Minsu Jang}, \bibinfo{person}{Youngwoo Yoon}, \bibinfo{person}{Jaewoo Choi}, \bibinfo{person}{Hyobin Ong}, {and} \bibinfo{person}{Jaehong Kim}.} \bibinfo{year}{2023}\natexlab{}.
\newblock \showarticletitle{A Structured Prompting based on Belief-Desire-Intention Model for Proactive and Explainable Task Planning}. In \bibinfo{booktitle}{\emph{Proceedings of the 11th International Conference on Human-Agent Interaction}} (Gothenburg, Sweden) \emph{(\bibinfo{series}{HAI '23})}. \bibinfo{publisher}{Association for Computing Machinery}, \bibinfo{address}{New York, NY, USA}, \bibinfo{pages}{375–377}.
\newblock
\showISBNx{9798400708244}
\urldef\tempurl%
\url{https://doi.org/10.1145/3623809.3623930}
\showDOI{\tempurl}


\bibitem[Jeromela and Conlan(2023)]%
        {jeromela2023onboarding}
\bibfield{author}{\bibinfo{person}{Jovan Jeromela} {and} \bibinfo{person}{Owen Conlan}.} \bibinfo{year}{2023}\natexlab{}.
\newblock \showarticletitle{Onboarding Stages and Scrutable Interaction: How Experts Envisioned Explainability in Proactive Time Management Assistants}. In \bibinfo{booktitle}{\emph{Proceedings of the 11th International Conference on Human-Agent Interaction}} (Gothenburg, Sweden) \emph{(\bibinfo{series}{HAI '23})}. \bibinfo{publisher}{Association for Computing Machinery}, \bibinfo{address}{New York, NY, USA}, \bibinfo{pages}{306–315}.
\newblock
\showISBNx{9798400708244}
\urldef\tempurl%
\url{https://doi.org/10.1145/3623809.3623851}
\showDOI{\tempurl}


\bibitem[Kenny and Huang(2024)]%
        {kenny2024utility}
\bibfield{author}{\bibinfo{person}{Eoin Kenny} {and} \bibinfo{person}{Weipeng Huang}.} \bibinfo{year}{2024}\natexlab{}.
\newblock \showarticletitle{The Utility of “Even if” Semifactual Explanation to Optimise Positive Outcomes}.
\newblock \bibinfo{journal}{\emph{Advances in Neural Information Processing Systems}}  \bibinfo{volume}{36} (\bibinfo{year}{2024}).
\newblock


\bibitem[Kenny and Keane(2021)]%
        {kenny2021generating}
\bibfield{author}{\bibinfo{person}{Eoin~M. Kenny} {and} \bibinfo{person}{Mark~T Keane}.} \bibinfo{year}{2021}\natexlab{}.
\newblock \showarticletitle{On Generating Plausible Counterfactual and Semi-Factual Explanations for Deep Learning}.
\newblock \bibinfo{journal}{\emph{Proceedings of the AAAI Conference on Artificial Intelligence}} \bibinfo{volume}{35}, \bibinfo{number}{13}, \bibinfo{pages}{11575--11585}.
\newblock
\urldef\tempurl%
\url{https://doi.org/10.1609/aaai.v35i13.17377}
\showDOI{\tempurl}


\bibitem[Maehigashi et~al\mbox{.}(2023)]%
        {maehigashi2023experimental}
\bibfield{author}{\bibinfo{person}{Akihiro Maehigashi}, \bibinfo{person}{Yosuke Fukuchi}, {and} \bibinfo{person}{Seiji Yamada}.} \bibinfo{year}{2023}\natexlab{}.
\newblock \showarticletitle{Experimental Investigation of Human Acceptance of AI Suggestions with Heatmap and Pointing-based XAI}. In \bibinfo{booktitle}{\emph{Proceedings of the 11th International Conference on Human-Agent Interaction}} (Gothenburg, Sweden) \emph{(\bibinfo{series}{HAI '23})}. \bibinfo{publisher}{Association for Computing Machinery}, \bibinfo{address}{New York, NY, USA}, \bibinfo{pages}{291–298}.
\newblock
\showISBNx{9798400708244}
\urldef\tempurl%
\url{https://doi.org/10.1145/3623809.3623834}
\showDOI{\tempurl}


\bibitem[McCloy and Byrne(2002)]%
        {mccloy2002semifactual}
\bibfield{author}{\bibinfo{person}{Rachel McCloy} {and} \bibinfo{person}{Ruth~M.J. Byrne}.} \bibinfo{year}{2002}\natexlab{}.
\newblock \showarticletitle{Semifactual “even if” thinking}.
\newblock \bibinfo{journal}{\emph{Thinking \& Reasoning}} \bibinfo{volume}{8}, \bibinfo{number}{1} (\bibinfo{year}{2002}), \bibinfo{pages}{41--67}.
\newblock
\urldef\tempurl%
\url{https://doi.org/10.1080/13546780143000125}
\showDOI{\tempurl}


\bibitem[Mnih et~al\mbox{.}(2013)]%
        {mnih2013playing}
\bibfield{author}{\bibinfo{person}{Volodymyr Mnih}, \bibinfo{person}{Koray Kavukcuoglu}, \bibinfo{person}{David Silver}, \bibinfo{person}{Alex Graves}, \bibinfo{person}{Ioannis Antonoglou}, \bibinfo{person}{Daan Wierstra}, {and} \bibinfo{person}{Martin Riedmiller}.} \bibinfo{year}{2013}\natexlab{}.
\newblock \showarticletitle{Playing atari with deep reinforcement learning}.
\newblock \bibinfo{journal}{\emph{arXiv preprint arXiv:1312.5602}} (\bibinfo{year}{2013}).
\newblock


\bibitem[Molnar(2020)]%
        {molnar2020interpretable}
\bibfield{author}{\bibinfo{person}{Christoph Molnar}.} \bibinfo{year}{2020}\natexlab{}.
\newblock \bibinfo{booktitle}{\emph{Interpretable machine learning}}.
\newblock \bibinfo{publisher}{Lulu. com}.
\newblock


\bibitem[Nugent et~al\mbox{.}(2009)]%
        {nugent2009gaining}
\bibfield{author}{\bibinfo{person}{Conor Nugent}, \bibinfo{person}{Dónal Doyle}, {and} \bibinfo{person}{Pádraig Cunningham}.} \bibinfo{year}{2009}\natexlab{}.
\newblock \showarticletitle{Gaining insight through case-based explanation}.
\newblock \bibinfo{journal}{\emph{Journal of Intelligent Information Systems}} \bibinfo{volume}{32}, \bibinfo{number}{3} (\bibinfo{year}{2009}), \bibinfo{pages}{267--295}.
\newblock
\showISSN{1573-7675}
\urldef\tempurl%
\url{https://doi.org/10.1007/s10844-008-0069-0}
\showDOI{\tempurl}


\bibitem[Olson et~al\mbox{.}(2021)]%
        {olson2021counterfactual}
\bibfield{author}{\bibinfo{person}{Matthew~L. Olson}, \bibinfo{person}{Roli Khanna}, \bibinfo{person}{Lawrence Neal}, \bibinfo{person}{Fuxin Li}, {and} \bibinfo{person}{Weng-Keen Wong}.} \bibinfo{year}{2021}\natexlab{}.
\newblock \showarticletitle{Counterfactual state explanations for reinforcement learning agents via generative deep learning}.
\newblock \bibinfo{journal}{\emph{Artificial Intelligence}}  \bibinfo{volume}{295} (\bibinfo{year}{2021}), \bibinfo{pages}{103455}.
\newblock
\showISSN{0004-3702}
\urldef\tempurl%
\url{https://doi.org/10.1016/j.artint.2021.103455}
\showDOI{\tempurl}


\bibitem[Puiutta and Veith(2020)]%
        {puiutta2020explainable}
\bibfield{author}{\bibinfo{person}{Erika Puiutta} {and} \bibinfo{person}{Eric~MSP Veith}.} \bibinfo{year}{2020}\natexlab{}.
\newblock \showarticletitle{Explainable reinforcement learning: A survey}. In \bibinfo{booktitle}{\emph{International cross-domain conference for machine learning and knowledge extraction}}. Springer, \bibinfo{pages}{77--95}.
\newblock


\bibitem[Rosenfeld and Richardson(2019)]%
        {rosenfeld2019explainability}
\bibfield{author}{\bibinfo{person}{Avi Rosenfeld} {and} \bibinfo{person}{Ariella Richardson}.} \bibinfo{year}{2019}\natexlab{}.
\newblock \showarticletitle{Explainability in human–agent systems}.
\newblock \bibinfo{journal}{\emph{Autonomous Agents and Multi-Agent Systems}} \bibinfo{volume}{33}, \bibinfo{number}{6} (\bibinfo{year}{2019}), \bibinfo{pages}{673--705}.
\newblock
\showISSN{1573-7454}
\urldef\tempurl%
\url{https://doi.org/10.1007/s10458-019-09408-y}
\showDOI{\tempurl}


\bibitem[Sutton and Barto(2018)]%
        {sutton20018reinforcement}
\bibfield{author}{\bibinfo{person}{Richard~S Sutton} {and} \bibinfo{person}{Andrew~G Barto}.} \bibinfo{year}{2018}\natexlab{}.
\newblock \bibinfo{booktitle}{\emph{Reinforcement learning: An introduction}}.
\newblock \bibinfo{publisher}{MIT press}.
\newblock
\showISBNx{0262352702}


\bibitem[Tsirtsis et~al\mbox{.}(2021)]%
        {tsirtsis2021counterfactual}
\bibfield{author}{\bibinfo{person}{Stratis Tsirtsis}, \bibinfo{person}{Abir De}, {and} \bibinfo{person}{Manuel Rodriguez}.} \bibinfo{year}{2021}\natexlab{}.
\newblock \showarticletitle{Counterfactual Explanations in Sequential Decision Making Under Uncertainty}. In \bibinfo{booktitle}{\emph{Advances in Neural Information Processing Systems}}, \bibfield{editor}{\bibinfo{person}{M.~Ranzato}, \bibinfo{person}{A.~Beygelzimer}, \bibinfo{person}{Y.~Dauphin}, \bibinfo{person}{P.S. Liang}, {and} \bibinfo{person}{J.~Wortman Vaughan}} (Eds.), Vol.~\bibinfo{volume}{34}. \bibinfo{publisher}{Curran Associates, Inc.}, \bibinfo{pages}{30127--30139}.
\newblock
\urldef\tempurl%
\url{https://proceedings.neurips.cc/paper_files/paper/2021/file/fd0a5a5e367a0955d81278062ef37429-Paper.pdf}
\showURL{%
\tempurl}


\bibitem[Tsirtsis and Rodriguez(2023)]%
        {tsirtsis2023finding}
\bibfield{author}{\bibinfo{person}{Stratis Tsirtsis} {and} \bibinfo{person}{Manuel Rodriguez}.} \bibinfo{year}{2023}\natexlab{}.
\newblock \showarticletitle{Finding Counterfactually Optimal Action Sequences in Continuous State Spaces}. In \bibinfo{booktitle}{\emph{Advances in Neural Information Processing Systems}}, \bibfield{editor}{\bibinfo{person}{A.~Oh}, \bibinfo{person}{T.~Naumann}, \bibinfo{person}{A.~Globerson}, \bibinfo{person}{K.~Saenko}, \bibinfo{person}{M.~Hardt}, {and} \bibinfo{person}{S.~Levine}} (Eds.), Vol.~\bibinfo{volume}{36}. \bibinfo{publisher}{Curran Associates, Inc.}, \bibinfo{pages}{3220--3247}.
\newblock
\urldef\tempurl%
\url{https://proceedings.neurips.cc/paper_files/paper/2023/file/09ae6beae5f1ff38f05c05979097ea0f-Paper-Conference.pdf}
\showURL{%
\tempurl}


\bibitem[Verma et~al\mbox{.}(2020)]%
        {verma2020counterfactual}
\bibfield{author}{\bibinfo{person}{Sahil Verma}, \bibinfo{person}{John Dickerson}, {and} \bibinfo{person}{Hines Keegan}.} \bibinfo{year}{2020}\natexlab{}.
\newblock \bibinfo{title}{Counterfactual explanations for machine learning: A review}.
\newblock
\newblock


\bibitem[Wachter et~al\mbox{.}(2017)]%
        {wachter2017counterfactual}
\bibfield{author}{\bibinfo{person}{Sandra Wachter}, \bibinfo{person}{Brent Mittelstadt}, {and} \bibinfo{person}{Chris Russell}.} \bibinfo{year}{2017}\natexlab{}.
\newblock \showarticletitle{Counterfactual explanations without opening the black box: Automated decisions and the GDPR}.
\newblock \bibinfo{journal}{\emph{Harv. JL \& Tech.}}  \bibinfo{volume}{31} (\bibinfo{year}{2017}), \bibinfo{pages}{841}.
\newblock


\end{thebibliography}

\end{document}